# Cross-lingual Transfer of Abstractive Summarizer to Less-resource Language

Aleš Žagar · Marko Robnik-Šikonja

**Abstract** Automatic text summarization extracts important information from texts and presents the information in the form of a summary. Abstractive summarization approaches progressed significantly by switching to deep neural networks, but results are not yet satisfactory, especially for languages where large training sets do not exist. In several natural language processing tasks, a cross-lingual model transfer is successfully applied in less-resource languages. For summarization, the cross-lingual model transfer was not attempted due to a non-reusable decoder side of neural models that cannot correct target language generation. In our work, we use a pre-trained English summarization model based on deep neural networks and sequence-to-sequence architecture to summarize Slovene news articles. We address the problem of inadequate decoder by using an additional language model for the evaluation of the generated text in target language. We test several cross-lingual summarization models with different amounts of target data for fine-tuning. We assess the models with automatic evaluation measures and conduct a small-scale human evaluation. Automatic evaluation shows that the summaries of our best cross-lingual model are useful and of quality similar to the model trained only in the target language. Human evaluation shows that our best model generates summaries with high accuracy and acceptable readability. However, similar to other abstractive models, our models are not perfect and may occasionally produce misleading or absurd content.

A. Žagar and M. Robnik-Šikonja
University of Ljubljana, Faculty of Computer and Information Science, Večna pot 113, Ljubljana, Slovenia.
E-mail: ales.zagar@fri.uni-lj.si, E-mail: marko.robnik@fri.uni-lj.si

# 1 Introduction

Summarization is a process of extracting or collecting important information from texts and presenting that information in the form of a summary. According to the output of the process, summarization can be broadly divided into an extractive and abstractive type. The extractive approach is non-productive in a sense that it copies important sentences, and the resulting summary does not include new words or sentences. The abstractive approach is creative and produces summaries that rephrase the given content and can contain originally unused words.

The abstractive neural summarization approaches use similar deep learning architectures as machine translation (MT), but face some additional problems: the input is usually longer, the output is short compared to the input, and the content compression is lossy. Current abstractive summarization may suffer from repetitive outputs (n-gram repetition), absurd content (creating meaningless sentences and phrases), mis-represented facts (e.g., who won the football match), problems with out-of-vocabulary words (applies to models without a copy mechanism which omit many proper names), or poor content selection (especially for longer texts). Nevertheless, the returned summaries are often useful and of good quality.

Many summarization approaches exist for resource-rich languages [2, 6, 40]. Existing cross-lingual approaches address the problem of a document in one language and its summary in another language, typically English or Chinese [50, 33], while we are interested in the cross-lingual transfer of trained summarization models from resource rich-languages to less-resourced languages, i.e. to produce summaries in a less-resourced language. In classification, cross-lingual embeddings present a promising approach for less-resource languages and enable the model transfer from resource-rich to less-resourced languages [1, 3, 26]. Typically, this is done by multilingual models such as BERT [13], or training the model on the resource-rich language (using monolingual embeddings in the source language) and then applying it to the less-resourced language where the input embeddings in the target language are mapped to the source language embeddings. Unfortunately, this standard procedure does not work for cross-lingual summarization, as the model is trained to output the sentences in the grammar of the source language. Blindly applying the procedure to a summarization model trained on English would produce sentences with English grammar in the target language. It is possible to achieve cross-lingual summarization using translation, but for summarization, this approach is unsatisfactory, as our baseline models, described in Section 5.1 show.

In the proposed solution, we use a pretrained English summarization model, proposed by Chen and Bansal [9], and use English as the source language and Slovene as the less-resourced target language. Using cross-lingual embeddings, we map Slovene word embeddings into the English word vector space. As zero-shot transfer learning is not satisfactory, we further fine-tune the resulting model. Our cross-lingual models are trained with increasingly large portions of the available target language dataset. In the output stage of our models, we generate several hypotheses and selected the best one using four evaluation metrics, including a transformer-based neural language model in the target language.

Our main contribution is the cross-lingual methodology that produces a useful summarization model for a less-resourced language. The automatic metrics show that the created summarizer is on par with a summarization model trained from scratch on the target language. In a zero-shot transfer, our cross-lingual approach does not require any resources in the target language apart from a monolingual corpus to build a language model. In a few-shot transfer, a moderate amount of summaries in the target language greatly improve the outputs.

The paper is split into further five sections. In Section 2, we present related works, and in Section 3, we describe the Slovene datasets to build the output selection language models and to fine-tune the summarization model in the few-shot transfer experiments. Section 4 outlines the proposed cross-lingual summarization model and gives details of the used components. We report the results in Section 5, and present conclusions and ideas for further work in Section 6.

## 2 Background and related work

We split this section into three parts. In Section 2.1, we first describe monolingual approaches to text summarization in English and other languages, followed by cross-lingual summarization attempts in Section 2.2. As our approach is based on cross-lingual embeddings, we shortly outline relevant background in Section 2.3.

2.1 Monolingual approaches to text summarization

Most early summarization approaches used the extractive approach also suitable for a multi-document summarization [16]. Lately, deep neural networks learning sequence to sequence (seq2seq) transformations produced state of the art abstractive summaries [39, 31]. Seq2seq models first encode a source document into an internal numeric representation and then decode it into an abstractive summary. These models work best for short single-document summaries, e.g., headline generation and news summarization. They use the attention mechanism which ensures that the decoder focuses on the appropriate input words [5]. They frequently use the copy mechanism that copies relevant words from the input when they are not present in a dictionary [41], and the coverage mechanism that avoids redundant contents [44]. Auxiliary tasks, e.g., keyphrase extraction, can improve the summarization results [27]. Currently, all of the best summarization models [37], [48], [14] are based on the transformer architecure [45].

As we use Slovene as the target language, we report the work on summarization in this language. Recently, Zidarn [51] built the first abstractive summarizer for the Slovene language using the seq2seq architecture and deep neural networks. The best results were produced by a two-layer LSTM with attention mechanism, copy mechanism, and beam search. To allow comparison, we use the same target language dataset of approximately 120,000 news. Zidarn [51] showed that this dataset is not large enough to achieve results comparable to English.

Besides English, there are only a few other languages with abstractive summarizers. Straka et al. [42] presented SumeCzech, a large news summarization dataset for Czech (1 million samples). For summarization, they compared unsupervised methods such as TextRank [28], returning a few first sentences, and supervised methods (logistic regression and random forests) on handcrafted features. Fecht et al. [15] used the encoder-decoder architecture on German Wikipedia articles (100,000 samples), where the summary is the first section of the article and the subsequent text represents the document. Hu et al. [20] created a Chinese summarization dataset (2.4 million samples) from a Chinese microblogging website Sina Weibo and used a recurrent neural network for abstractive summarization.

2.2 Cross-lingual approaches to text summarization

Most existing cross-lingual summarization attempts aim to obtain a summary in a different language than the original document. For that purpose, they use summarization in combination with MT. Zhu et al. [50] proposed a cross-lingual approach suitable for resource-rich languages where both source and target language have enough training data to build a summarizer. Two different translation schemes are used: "translate then summarize" scheme first translates the original document into the target language and then generates a summary; "summarize then translate" scheme first generates a summary and then translates it into the target language. Zhu et al. [50] used the MT on a large English and Chineese corpus to first create a cross-lingual summarization dataset and then trained a neural network in an end-to-end manner incorporating both MT and summarization.

Ouyang et al. [33] aimed at summarizing documents in low-resource languages in the resource-rich language (English). To address the problem of noisy MT from low-resource languages, they translated documents from an English document-summary corpus to three low-resource languages and back into English. They coupled noisy documents with the original summaries and trained the neural network summarization architecture proposed by See et al. [41] on the obtained corpus. The approach was shown to improve over the "translate then summarize" scheme as the neural network took into account some of the errors introduced by MT from less-resourced languages. In our work, we address a situation where we want to obtain the summary in the same less-resource language as the original text. Our cross-lingual approach uses the pretrained summarization model in the resource-rich language and fine-tunes it to the target language. We use MT as a baseline (translate-summarize-translate), and show that it is not competitive with our direct cross-lingual model transfer approach.

Chi et al. [10] outperformed machine-translation-based approaches in a headline generation task by pre-training a seq2seq transformer model [45] under both monolingual and cross-lingual settings. For the pretraining procedure, they used various tasks: monolingual masked language modeling, denoising auto-encoding objective (to pre-train the encoder-decoder attention mechanism), cross-lingual masked language modeling, and cross-lingual auto-encoding. After the pretraining phase, the model was fine-tuned on question generation and abstractive summarization tasks. In contrast to the headline generation task, where the outputs are short and require little grammar, we work with much longer summaries. To accommodate to less-resourced languages, our approach uses cross-lingual word embeddings at the input to the al-ready pretrained summarization model and adapts the decoder phase to fit the target language better.

Our cross-lingual approach is based on the monolingual model proposed by Chen and Bansal [9]. This hybrid summarization model first selects salient sentences and then paraphrases them. The model is comprised of two independently trained neural networks bridged by policy-based reinforcement learning. We describe this model in Section 4.2.

## 2.3 Word embeddings

The idea of word embeddings is to learn high-dimensional vectors that capture the meaning of words. Popular variants are Word2vec [29], GloVe [35], fastText [18], ELMo [36], and BERT [13]. An important insight for our work is that relations between words in the embedded space are preserved across languages [30]. Cross-lingual embeddings align monolingual embeddings into a joint vector space [38]. In the beginning, these techniques required parallel corpora or a bilingual dictionary to map words from a source to target language. Recent approaches can train cross-lingual embeddings in an unsupervised manner [23, 3]. A major drawback of classical word embeddings is that they cannot deal with polysemy. Recent contextual embeddings, ELMo [36] and BERT [13], learn a different representation for each word based on its context.

## 3 Datasets

We describe the creation of two datasets, one for the summarization task and the other for the language modeling used in the output selection. Both datasets were extracted from the Gigafida corpus [21] of written standard Slovene. The corpus consists of newspapers, magazines, and web texts, and contains 38,310 documents with more than 1.1 billion words. We end the section with a short discussion on approximations to true human summaries used in existing summarization datasets.

## 3.1 Slovene summarization dataset

The summarization dataset contains news and their summaries from the Slovenian press agency (STA) news web texts. The first paragraph of each news article is taken as a summary and the rest of it as the text of the news. Since the Gigafida corpus from which we extracted STA news is sentence segmented but not paragraph segmented, we designed a heuristic to extract the first paragraph. We started with 284,000 training samples but kept only texts between 1,000 and 3,000 characters. Some texts were discarded as they contained weather reports, lists of events around the world, etc., and some of them were too long. A total of 127,563 samples remained and were split into the train, test and validation set. Both the test and validation set contain 5,000 instances, and the training set contains the remaining 117,563 news.

### 3.2 Slovene language model dataset

To create our cross-lingual summarization model, we started with the trained English model. While a cross-lingual mapping can transfer the target language (Slovene) into the required input space of the source language (English), this is not sufficient to produce sensible texts in the target language because the grammar of the decoder remains in the source language. Our output modifications require that we train a language model in the target language. For that purpose, we trained a character-level Slovene language model. Bojanowski et al. [7] discovered that language models for morphologically rich languages (such as Slovene) are improved by using character-level information. As the training set we used the Gigafida corpus which is tokenized and sentence segmented. All punctuation, special characters, and numbers were preserved, but alphabetical characters were lower-cased. A total of 59,861,870 sentences were extracted with the average sentence length of 242 characters. The sentences were split into the train, test, and validation set with ratios of 90:5:5.

### 3.3 Approximations to true human summaries

The aim of our work is to produce methodology for cross-lingual transfer of trained summarizers. To evaluate such a system in zero-shot and few-shot transfer mode, we need a reasonably sized dataset in the target less-resourced language. Unfortunately, there is no such summarization dataset with actual human abstracts in Slovene and our investigation showed that the same is true for other languages, as all existing large datasets use approximations.

The most commonly used English summarization dataset CNN/DM [31] does not contain actual human abstracts but only the main bullet points (highlights). Another widely used English dataset, the Gigaword summarization dataset [17], is a headline generation task. The Newsroom summaries were produced from the metadata available in the HTML pages of articles using various keywords with no standard metadata format [19]. Non-English datasets are produced in a similar way. For example, in the Slovak SME dataset, Suppa and Adamec [43] joined the headline of an article with its lead paragraph to form the target summary.

We could not find any real abstracts of the described datasets for human evaluation purposes. The only datasets we are aware of and contain proper abstracts are too small and not appropriate for neural summarization [24, 34]. If we built a new small Slovene dataset with the actual abstracts, it would not be sufficiently large for training and would not match the properties of the training datasets. Such a small new dataset would introduce a task transfer problem and we would lose the possibility to compare our results with other approaches (e.g., Slovene evaluation in Table 7), which used existing approximating datasets. We believe that this task transfer should be approached in further work and studied carefully.

Large English datasets which do contain actual abstracts are based on much longer texts, e.g., ArXiv and PubMed abstracts [11] or book summaries [22]. These datasets are typically not treated with neural abstractive summarization approaches used in our work but use an extractive approach or a hybrid extractive-abstractive approach. These approaches are outside the scope of this work.

## 4 Architecture and implementation of cross-lingual summarizer

In this section, we first outline our solution to the problem of cross-lingual summarization. After that, we provide descriptions of components used: cross-lingual word embeddings for the input, fine-tuning of pretrained English summarization model to Slovene, generation and evaluation of the best hypothesis with several evaluation scores, including the Slovene language model.

### 4.1 Architecture of the cross-lingual summarizer

The proposed approach consists of several steps, presented in Figure 1. Below we describe them step-by-step.

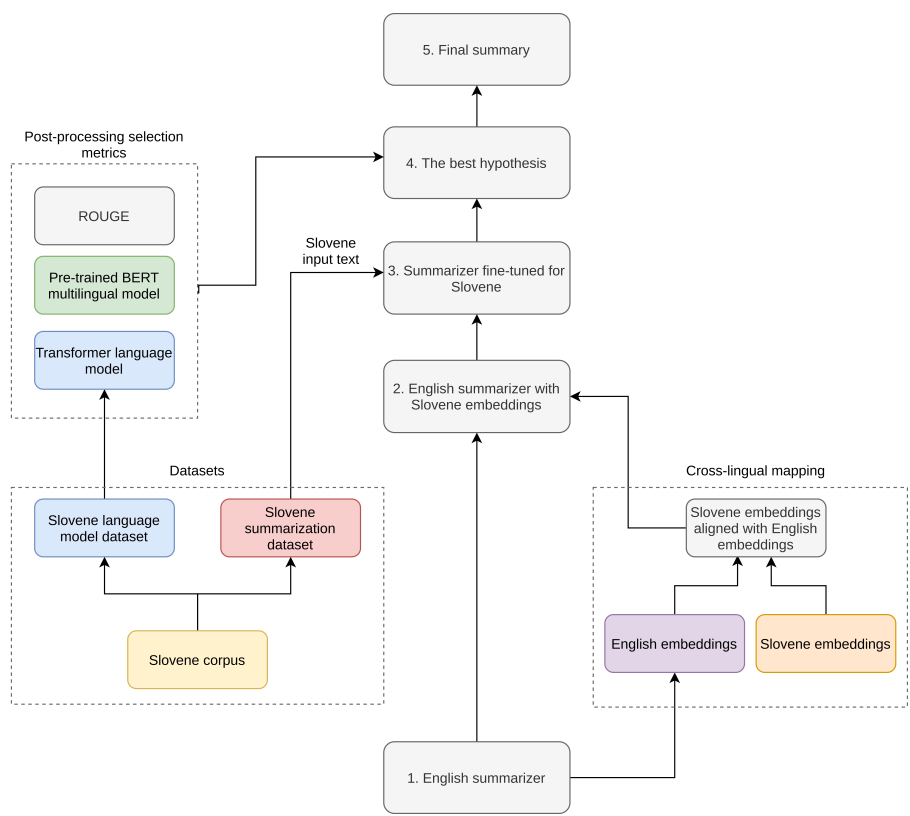

**Fig. 1** The outline of the proposed cross-lingual summarization approach.

As a pretrained summarization engine (step 1), we could use several pretrained summarization models, but in this work, we used the English summarizer [9], as described in Section 4.2. To adapt it to cross-lingual setting, we first replaced the English word embeddings at its input with Slovene embeddings (step 2), as described in Section 4.3. To match the word semantics of the two languages, we used the cross-lingual Procrustes alignment [23] and mapped the Slovene word embeddings into the English vector space. This already allows us to put Slovene text on the input of the summarization model (step 3). We fine-tuned the model with different amounts of Slovene text as discussed in Section 4.4. In step 4, we used the trained model to generate several hypotheses, and in step 5, we assessed the hypotheses to choose the final output. This assessment used an independently trained Slovene language model using transformer architecture (described in Section 4.5) and two different metrics, described in Section 4.6. The best hypothesis was included into a summary.

### 4.2 English summarization model

As our source language summarization model, we used the pretrained summarization model proposed by Chen and Bansal [9]. The model uses customarily trained word2vec embeddings and thus allows a cross-lingual mapping. The architecture of the model is relatively complex and belongs to hybrid approaches to text summarization that combine abstractive and extractive elements. On a high level, it consists of i) the extractive network (that selects salient sentences), ii) the abstractive network (that rewrites or paraphrases them), and iii) the reinforcement learning (RL) step that optimizes the model end-to-end. Both the extractor and abstractor networks are trained independently. During the RL step, the model updates only the extractor weights and leaves the abstractor as it is. The model was trained on the CNN/DailyMail dataset[1], which contains 287,226 training summary/text pairs, 13,368 validation pairs and 11,490 test pairs. The details are available in [9].

### 4.3 Cross-lingual input alignment

At the input to the neural network summarization model, words are encoded into numeric vectors using word embeddings. In our cross-lingual setting, we use the Slovene input and map it into the English vector space. As the Slovene embedding model, we used the pretrained Slovene fastText embeddings [18], trained on a mixture of Slovene Wikipedia and Common Crawl data[2]. FastText embeddings are constructed with the word2vec CBOW algorithm [29], extended with position weights and subword information. FastText embeddings are especially suitable for morphologically rich languages such as Slovene. To transform Slovene embeddings into the English vector space, we used the MUSE library [23] in a supervised setting. For this transformation, MUSE internally created a train dictionary of size 5,000 and a test dictionary of size 1,500. We replaced the English dictionary with the Slovene dictionary which was built from 30,000 most common words in the Slovene training dataset. The role of the dictionary is to map words to their embeddings.

---

[1] https://cs.nyu.edu/ kcho/DMQA/
[2] https://fasttext.cc/docs/en/crawl-vectors.html

### 4.4 Fine-tuning of the summarization model

Once the target language input (Slovene) is mapped to the source language (English), it is used as an input to the summarization model. Such a model can be already used for summarization in the target language (zero-shot transfer). The resulting summaries adhere to the source language grammar and are of low quality. If any target language summaries are available, we can improve the summarization model with additional training instances (few-shot transfer). To analyze the required amount of additional data, we created several models, presented in Table 1. The models differ in the quantity of additional target language data used in their fine-tuning.

| Model | Slovene dataset size in % | # instances | Details |
|---|---|---|---|
| MENG | 0% | 0 | cross-lingual mappings, no fine-tuning, zero-shot transfer |
| M1 | 1% | 1,176 | cross-lingual mappings, trained extractor, fine-tuned abstractor |
| M10 | 10% | 11,756 | cross-lingual mappings, trained extractor, fine-tuned abstractor |
| M25 | 25% | 29,391 | cross-lingual mappings, trained extractor, fine-tuned abstractor |
| M50 | 50% | 58,782 | cross-lingual mappings, trained extractor, fine-tuned abstractor |
| M100 | 100% | 117,563 | cross-lingual mappings, trained extractor, fine-tuned abstractor |
| MSLO | 100% | 117,563 | Slovene embeddings, trained extractor, trained abstractor, no transfer |

**Table 1** The produced models using different amounts of target language data (Slovene) in the fine-tuning of the original summarization model.

MENG is the baseline zero-shot transfer model, which means that no target language data was used, only the English embeddings were swapped with the mapped Slovene embeddings. The models M1, M10, M25, M50, and M100 use 1%, 10%, 25%, 50%, or 100% of our target language training set (see Section 3) to fine-tune the English model. We also trained the extractor part of each model because only the reinforcement learning optimized extractor was provided by Chen and Bansal [9]. Simultaneously, we updated the weights of the pretrained abstractor and, in the final step, optimized the models with the RL component.

MSLO is not a cross-lingual model and was trained on the complete target language training set from scratch. Note that the training set in the target language is significantly smaller than the training set in the source language (117,563 summaries for MSLO vs 287,226 for MENG).

### 4.5 Training the Slovene language model

The adapted and fine-tuned models produce summaries in Slovene, but the quality is not always adequate. For that reason, we used the decoder to generate several hypotheses, post-processed them, and selected the best one according to different evaluation approaches (described below in Section 4.6). As we aim to optimize the fluency and grammatical correctness of the output sentences, one of the evaluation approaches uses language models. For that purpose, we trained a character-level language model in the target language (Slovene).

Many of the current state-of-the-art language models [4, 12], trained on datasets similar to ours [8], use variants of the transformer architecture [45]. We used the transformer decoder as implemented in the Tensor2tensor library [46], Adam optimizer, 8 attention heads, 6 hidden layers, and position-wise feed-forward networks with one hidden layer of size 2048 and ReLU activation function. These are standard hyperparameters for training on a single GPU. We increased the maximum size of the input from 256 to 512, which is approximately the 95th percentile of the sentence character length in the training corpus. Shorter sentences are padded with spaces and longer are cut off. The dictionary contains 581 characters. The total number of learning parameters was 19,035,136.

The language model was trained for 100 epochs in two parts due to limited computational resources: 60 epochs using 30 million sentences, and 40 epoch with another 23 million sentences). The batch size was 2048. The model was evaluated on the test set with 10k sentences (see Section 3). Training took approximately 4 days on Nvidia Titan X 12GB GPU.

4.6 Creation of the final summary

The English summarization model is fine-tuned to produce Slovene summaries. Nevertheless, the outputs are sometimes of low quality. For example, sometimes summarization models produce repeating n-grams, which we eliminate with a rule-based approach. To improve the quality of summaries, we extracted a large number of hypotheses from the abstractive network and assessed them with different heuristics. In the search for hypotheses, we expanded the beam size from standard 4-16 to 64. The heuristic for the assessment of hypotheses consists of two components that try to capture the presence of relevant contents and the readability of hypotheses.

**Relevant content.** The quality of the content is assessed with two scores. ROUGE score is the standard metric for summarization quality [25] and uses weighted number of matching n-grams between the refrence summary and hypothesis. Recently proposed BERTScore [49] is based on the similarity of sentence representation with the pretrained multilingual BERT model [13]. We calculated the ROUGE and BERTScore scores by comparing the generated hypotheses from the abstractor network with the sentences extracted with the extractive network.

**Text readability.** The readability of the generated hypotheses is assessed with two measures: the internal evaluation of hypotheses with the loss function computed by the abstractive neural network, and the Slovene language model described in Section 4.5. The latter is expressed with the perplexity score, computed as the average entropy per character expressed in bits.

With this approach, we get four different assessments for each generated hypothesis. We first used only one heuristic to select the best hypothesis and analyzed the results. After that, we considered combinations of two heuristics. For example, we first used the ROUGE scores to narrow down the selection to 32 best hypotheses. These 32 hypotheses were scored anew by the language model and the best one according to the language model scores was selected. In combinations of two metrics, we did not require that they belong to different categories, i.e. we allowed a combination of two content-based heuristics or two-readability-based heuristics.

## 5 Evaluation

In this section, we provide the results and analyses of the created summarization models. We start with the presentation of the evaluation metrics and baseline models in Section 5.1. The results of baseline and fine-tuned models are presented in Section 5.2. The best of the fine-tuned models is further analyzed in Section 5.3 where we compare the proposed heuristics for the selection of output sentences. Section 5.4 contains the human evaluation of the best-produced model. We compare our results with the related approaches in Section 5.5. Finally, in Section 5.6, we manually analyze strengths and weaknesses of our best model.

### 5.1 Evaluation metrics and baselines

We first present the standard evaluation metrics used in summarization, ROUGE-1, ROUGE-2, and ROUGE-L. Next, we describe the baseline models, both monolingual and translation-based.

ROUGE (Recall-Oriented Understudy for Gisting Evaluation) scores are the most commonly used metrics in the evaluation of automatically generated text summaries [25]. It measures the quality of a summary based on the number of overlapping units (n-grams, sequences of texts, etc.) between reference summaries (created by humans) and automatically generated summaries. The most commonly used metrics are ROUGE-N and ROUGE-L. ROUGE-N measures the overlapping of n-grams, e.g., ROUGE-1 for unigrams and ROUGE-2 for bigrams. ROUGE-L measures the longest common subsequence found in the compared summaries.

As baseline summarization models, we use monolingual models and translation-based models. MSLO is a monolingual model, trained on the complete Slovene STA dataset. EXT Baseline is a purely extractive model that is part of the MSLO model. The third baseline monolingual model is PG, an end-to-end abstractive model [41]. PG is a hybrid between a seq2seq attention model based on LSTMs and pointer networks [47] that enable the model either to copy words via pointing or generate them from a fixed vocabulary. This helps to solve the problem of out-of-vocabulary words. PG uses a coverage mechanism to mitigate the problem of repetition of seq2seq models by preventing the model to focus on the same locations all the time.

To establish the MT baseline, we used the Google MT service. We translated the test set from Slovene into English and generated English summaries with the pretrained monolingual English summarizer. After that, we translated the generated English summaries back into Slovene.

5.2 Results of cross-lingual fine-tuning

As described in Section 4.4, the pretrained summarization model can be improved with different amounts of training data in the target language. Table 2 shows the results of the six models listed in Table 1 and the three baseline monolingual models, described in Section 5.1.

|  | Average generated | | Evaluation scores | | | |
|---|---|---|---|---|---|---|
| Model | sentences | characters | ROUGE-1 | ROUGE-2 | ROUGE-L | Perplexity |
| MENG | 3,99 | 500,61 | 18,91 | 3,74 | 16,27 | 3,69 |
| M1 | 2,81 | 218,48 | 12,94 | 1,96 | 11,61 | 4,23 |
| M10 | 1,95 | 204,59 | 15,71 | 3,71 | 13,87 | 2,14 |
| M25 | 2,89 | 159,00 | 19,32 | 5,00 | 17,12 | 2,19 |
| M50 | 3,01 | 168.59 | 21,30 | 6,09 | 18,91 | 2,15 |
| M100 | 2,79 | 297,67 | 21,67 | 6,81 | 19,16 | **2,12** |
| MSLO | 2,58 | 270,79 | 21,07 | 6,62 | 18,64 | 2,13 |
| MT Baseline | 4,02 | 297,06 | 19,76 | 3,64 | 17,14 | 4,26 |
| EXT Baseline | 2,58 | 510,37 | 22,71 | 5,58 | 18,46 | / |
| PG [41] | 1,79 | 270,73 | **23,57** | **7,76** | **20,04** | 3,15 |
| Reference Slovene | 2,10 | 302,02 | | | | |
| Reference English | 3,88 | 312,51 | | | | |

**Table 2** The performance of the cross-lingual models with different amounts of target language data (MENG, M1, M10, M25, M50, and M100) and the monolingual models (MSLO, MT Baseline, EXT Baseline, PG). The last two rows represent the statistics of reference Slovene and English summaries. We cannot compute the perplexity of the EXT model as this purely extractive model outputs human-written sentences.

The English monolingual model MENG generates more than twice as many character as the other models and on average 4 sentences while the other models generate 2 to 3. These numbers are the result of learning, as the dataset of English summaries contains on average more and longer summaries. The M1 model shows that as little as 1k of additional instances is enough to update the number of extracted sentences.

The metrics ROUGE-1, ROUGE-2, and ROUGE-L show similar relations between the compared models. Surprisingly, the zero-shot transfer model MENG scores higher on ROUGE metrics than M1 and M10. The reason for this is that it extracts more sentences, generates longer summary sentences, and repeats the sentences. Analyzing the results of MENG, we noticed that the model sometimes cannot finish a sentence properly, e.g., it generates good content, but does not stop and just continues to generate words. We speculate that the problem is in special tokens (start of the sentence, end of the sentence, etc.) that capture the grammar of source language. These special tokens may be a hidden problem in the cross-lingual seq2seq model transfer.

We manually inspected the returned summaries to assess their readability. M1 does not show any significant readability improvement over MENG, while M10 shows some improvement. MENG often generates long sentences with redundant and rare words, and inserts punctuation at inappropriate places. On the contrary, M1 generates too short sentences and summaries with many missing words. M10 shows an improvement in sentence selection over M1, and improvement in readability over both M1 and MENG. Still, most of the sentences are not well-formulated, but the meaning is present in almost all of them.

Models M25 and M50 show interesting properties, considering that they produce scores quite close to the models trained on much larger training sets in the target language (i.e. M100 and MSLO). This indicates utility of cross-lingual transfer which can produce useful models with significantly less data.

The PG model scores highest on ROUGE scores, but its perplexity is not on par even with M10. The reason for this is the pure abstractive nature of this model (other models are hybrid extractive-abstractive models). In the generation phase, the abstractive models are not constraint when choosing the content. The manual inspection shows that the PG model generates summaries with higher readability than MT Baseline but much lower than the cross-lingual models trained on sufficient amounts of data.

M100 (cross-lingual model) and MSLO (trained from scratch) are the best models for the Slovene summarization. These two models use the same amounts of training data. With manual inspection, we were unable to conclude which model is better in terms of readability. However, M100 consistently shows better ROUGE scores: ROUGE-1 is improved for 0.60, ROUGE-2 for 0.19, and ROUGE-L for 0.52. This shows that our cross-lingual approach produces better summaries compared to monolingual models even without additional sentence selection mechanism analyzed in Section 5.3.

5.3 Selection of the final output sentences

As explained in Section 4.6, we use our best cross-lingual summarization mode to generate 64 hypotheses for each of the extracted sentences. The candidate sentences are assessed with four heuristics (ROUGE-L, BERTscore, the internal loss value, and perplexity of the language model) and the best is included in the final summary. Table 3 shows the results.

As the baseline result, we report the scores of our best fine-tuned model M100 (taken from Section 5.2), which uses only the internal loss score to select the final output. All the selection heuristics improve the performance of the baseline model. We tested all combinations of the four selection metrics but report only the best one (in the last row of Table 3).

| Selection heuristics | ROUGE-1 | ROUGE-2 | ROUGE-L |
|---|---|---|---|
| M100 with no additional selection | 21,67 | 6,81 | 19,16 |
| M100 + Transformer LM | 22,53 | 6,83 | 19,61 |
| M100 + Multilingual BERTScore | 24,87 | 7,41 | 21,36 |
| M100 + ROUGE-L | 24,88 | 7,38 | 21,47 |
| M100 + ROUGE-L & BERTScore | **24,97** | **7,43** | **21,50** |

**Table 3** Selection of the output sentences from the hypotheses generated with the M100 model.

Initially, we hypothesized that two complementary metrics are needed to select the best hypothesis: one for the content and another for the readability. The last line of Table 3 shows that this is not the case: the best pair of heuristics consists of both content selection metrics, ROUGE-L and BERTscore. These results may be biased since the reported ROUGE metrics are content-based. The manual comparison of models with two complementary metrics and models with both content-based metrics confirmed that the former produced better readable summaries than the latter, but with lower content accuracy. We can conclude that the selection of output hypotheses significantly improves the quality of the output summaries.

5.4 Human evaluation

The automatic summary evaluation is limited in assessment of actual user needs and expectations [32]. For that reason, we organized a small study with human evaluation of generated summaries. For each full text, we used both the reference summary and the automatically generated candidate in a random order.

The task of referees was to assign the accuracy and readability score of a summary (see Table 4 for the scale of scores). The accuracy represents the amount of overlap between the given facts and the summarized information, and the readability measures fluency and comprehensibility of the summary. In our study, each of the 10 articles (two summaries per text, the generated and the reference) were evaluated by eight referees. Referees included three females and five males aged from 23 to 65, with different degrees of education.

| Score | Accuracy | Readability |
|-------|----------|-----------------|
| 1 | none | incomprehensible |
| 2 | little | poor |
| 3 | a lot of | acceptable |
| 4 | most of | good |
| 5 | all | flawless |

**Table 4** The scales for the accuracy and readability scores of summaries.

We report averages and standard deviations of the assigned scores in Table 5. Surprisingly, the accuracy of the reference summaries is lower than the accuracy of the generated summaries. We identified several reasons that explain this result. First, the reference summaries are actually the first paragraphs of news articles and often contain true facts and information that cannot be verified in the text. Unless misleading and speculative, the generated summaries produce verifiable content. Second, the evaluation scores do not directly measure the content quality of summaries. Following the instructions, participants may assign a high score to a summary that contains true but unimportant and irrelevant information. Third, our hybrid summarization model selects and paraphrases sentences. We assume that sometimes participants are lured into thinking that there is a greater content overlap between the text and the generated summary than between the text and a reference summary. Finally, our study is small and the standard deviation of the answers is considerable, therefore the results may be misleading. As anticipated, the readability score of the reference summaries is much higher than it is for the generated summaries.

| Type | Accuracy | Readability |
|---|---|---|
| Reference | 2,85 (1,24) | 4,18 (0,96) |
| Generated | 3,06 (1,18) | 3,41 (0,94) |

**Table 5** Average and standard deviation of human assigned accuracy and readability of reference and generated summaries.

## 5.5 Comparison with related research

We compare our best summarization model (M100 + ROUGE-L & BERTScore) to other existing summarization models for English (as an upper bound of existing technologies) and Slovene. Table 6 shows the results reported by authors of the listed models. In addition to the standard ROUGE scores, we also provide BERTscore where possible. The reported scores are not directly comparable but give a general picture of the success of the proposed cross-lingual approach.

| Model | ROUGE-1 | ROUGE-2 | ROUGE-L | BERTScore |
|---|---|---|---|---|
| Zidarn [51] Slovene LSTM | 23,77 | 7,97 | 23,95 | 0,679 |
| Our M100 + ROUGE-L & BERTScore | 24,97 | 7,43 | 21,50 | 0,679 |
| English Chen and Bansal [9] | 40,88 | 17,80 | 38,54 | \ |
| English Zhang et al. [48] - PEGASUS | 44,17 | 21,47 | 41,11 | \ |

**Table 6** Comparison of our best model with related Slovene model and state-of-the-art English models.

The only other neural summarization model for Slovene was built by Zidarn [51] who used a two-layer LSTM neural network with the attention mechanism, copy mechanism, and beam search. The dataset of this model is the same STA news dataset extracted from Gigafida corpus, but the author uses different train, test, and validation splits. Our model scored higher on ROUGE-1 (1.20 difference) but lower on ROUGE-2 (0.54) and ROUGE-L (2.45). The BERTScore results of both models are identical. Given the existing sources of variation (different subsets of the original data, different splits, and the problematic nature of automatic summary evaluation metrics), we can conclude that both models perform similarly.

Table 7 shows human evaluation of our best model and the best model of Zidarn [51]. For both models, human reported scores of the generated and reference summaries are presented. Both models produce acceptable readability scores, but in terms of accuracy, it seems that our model generates more accurate content.

| Model | Text type | Accuracy | Readability |
|---|---|---|---|
| Our M100 + ROUGE-L & BERTScore | Reference | 2,85 (1,24) | 4,18 (0,96) |
| Zidarn [51] Slovene LSTM | Reference | 2,61 (1,39) | 3,48 (1,04) |
| Our M100 + ROUGE-L & BERTScore | Generated | 3,06 (1,18) | 3,41 (0,94) |
| Zidarn [51] Slovene LSTM | Generated | 1,95 (1,24) | 3,10 (1,27) |

**Table 7** Average and standard deviation of human assigned accuracy and readability of reference and generated summaries.

As the bottom part of Table 6 shows, neither cross-lingual nor monolingual Slovene models can compare to English models in terms of performance. English models are usually trained either on the 4 million instances of the Gigaword dataset, appropriate for headline generation, or the 290k CNN/Daily Mail dataset, which is similar but larger than our Slovene dataset. The English model used in our experiments [9] achieves scores that are almost twice as high compared to our Slovene model. Its results are less misleading and mostly represents facts and information accurately. Many manually inspected summaries show that it omits less important dependent clauses. In our model, this behaviour is less frequent.

PEGASUS [48] is currently one of the best abstractive summarization models. It is based on the transformer neural architecture and presents an interesting novel insight: models are fine-tuned faster and more successfully if they are pretrained on tasks similar to the final task. Authors thus propose two pretraining objectives. One is the BERT masked language model known from [13]. Another is the gap sentence generation that selects and masks whole sentences from documents, and concatenates the gap-sentences into a pseudo-summary. The model is pretrained on two very large corpora. The C4 dataset consists of texts from 350M web-pages (750GB). The HugeNews dataset is even larger with 1,5B articles (3,8TB). The model achieved state of the art performance on 12 summarization tasks.

5.6 Manual analysis of strengths and weaknesses

In this section, we manually analyze three outputs of different quality from our best model (M100 + ROUGE-L & BERTScore) (Tables 8, 9, and 10). In the tables, "Slovene reference summary" represents the first paragraph of an article. The most important explanatory factor for the differences in quality seems to be the topic of a document. The model generates satisfactory summaries for texts with political and financial content, which represent the majority of the fine-tuning dataset (STA news). For the comprehensibility sake, we manually translated all the texts from Slovene to English, preserving the problems.

The first example in Table 8 demonstrates a good quality result. The summary is short, contains the essential information expressed with well-formulated sentences, and exhibits a certain level of abstraction. It replaces the phrase "the croatian news agency hina wrote that ... " with "foreign news agencies reported that ... " and cuts off the supplemntary information that starts with "announcing that austria would...". In the second sentence, the phrase "european council president donald tusk" is omitted for no apparent reason. The sentence uses a pronoun "they" for a replacement of the phrase "the austrian news agency apa", which indicates abstractive qualities.

The second example in Table 9 shows that the model can be misleading and factually inconsistent with the text. The mentioned play will not premiere in Ljubljana but in Maribor. The model speculates that the play will start at 8 p.m, although the text says thursday night. The third example in Table 10 shows that the model correctly identifies winners and loosers, but misrepresents the numbers (and some of the names), which was one of the most frequently observed errors.

| |
|---|
| **Human translation of the original Slovene article** |
| the croatian news agency hina reported that the slovenian government had expressed a negative opinion on the austrian control of the border with slovenia, announcing that austria would extend the control of the internal schengen border with slovenia for another six months. the austrian news agency apa also reported about it. hina also reported that slovenian prime minister marjan šarec will meet with the european council president donald tusk and the european commission president jean - claude juncker in an official visit to brussels in the autumn. the latter has recently been the subject of much criticism in ljubljana for its alleged bias in the arbitration dispute between slovenia and croatia. hina wrote that slovenian foreign minister miro cerar, currently visiting washington, expects relations between slovenia and the united states to improve. he intends to better inform the americans about the arbitration dispute between slovenia and croatia, as in his opinion us is not sufficiently acquainted with this problem. the serbian news agency tanjug reported that the slovenian police unions (the slovenian police union and the union of slovenian police), will resume strike activities on monday, which froze in march. tanjug also reported that the serbian president aleksander vučić received today the slovenian ambassador to serbia, vladimir gasparič, on a farewell visit. on this occasion, gasparic expressed his belief that the planning of the visit, which vucic and slovenian president borut pahor had recently discussed, was an additional incentive for good cooperation between the two countries. |
| **Human translation of the Slovene reference summary** |
| foreign news agencies wrote that the slovenian government had expressed a negative opinion on austrian control of the border with slovenia, announcing that austria would extend control of the internal schengen border with slovenia. they also reported that the slovenian police unions would resume preparations for strike activities. |
| **Human translation of the generated summary from Slovene**, ROUGE-L = 51,46 |
| foreign news agencies reported that the slovenian government had expressed a negative opinion on austrian control of the border with slovenia. they also reported that slovenian prime minister marjan šarec will meet with european commission president jean - claude juncker on an official visit to brussels in the autumn. |
| **The generated summary in Slovene** |
| tuje tiskovne agencije so poročale o tem , da je slovenska vlada izrazila negativno mnenje o avstrijskem nadzoru na meji s slovenijo . poročale so tudi , da se bo slovenski premier marjan šarec na jesenskem uradnem obisku v bruslju srečal s predsednikom evropske komisije jean - claudom junckerjem . |

**Table 8** The first example (good quality) of a summary produced by the best cross-lingual summarizer.

## 6 Conclusion and further work

We developed a neural cross-lingual approach to abstractive summarization. Our solution is based on the pretrained model in the resource-rich language (English), whose outputs are fine-tuned to the target language (Slovene) and further refined with sentence selection heuristics. We first showed that zero-shot transfer is unsatisfactory due to its output following the grammar of the source language. In few-shot transfer, we tested how different amounts of training data in target language used in fine-tuning affects the model and discovered that even small amounts of data in the target language significantly improve the quality of produced summaries. Nevertheless, the quantity and quality of the training sets play a huge role, and the target language dataset (Slovene) is not competitive in either respect. This is most evident when analyzing diverse topics from the Slovene dataset, where better-represented topics are better summarized compared to less represented ones. In addition to the automatic evaluation, we manually analyzed the quality of the results and also conducted a small-scale human evaluation. The assessments show that the accuracy and readability of the generated summaries are acceptable. Two additional contributions of our work are the first Slovene summarization dataset consisting of news articles, and publicly available character-based transformer neural language model. The source code of our system is freely available[3].

| |
|---|
| **Human translation of the original Slovene article** |
| the magnificent play of shadows and sound is in the hands of animators barbara jamšek and elene volpi. the show, which will premiere on thursday night, is based on a motif by dennis haseleye's picture book about a pirate trying to catch the moon and with songs from the žmavc press stage. "the story is about a greedy pirate who wants to steal the whole world, and in the end reaches for the moon," director tin grabnar told the news conference today. such a story, in his opinion, is an excellent starting point for a shadow theater performance, where the material world in the form of puppets and other props is placed in relation to the immaterial in the form of light and shadows. the performance takes place on a ship with two sails, set in a recently restored church, and serves both as a stage and a grandstand for spectators. the viewer is placed at the center of the action and, in the words of the author of the artistic image darko erdelja, has the feeling that he is at sea, "limited by matter, but by craving for more". the main language of the play is shadows, not words, as the text of the picture book has been severely curtailed in order to achieve a greater contrast between the material and the immaterial, according to the playwright katarina klančnik kocutar. "appropriating everything material, but at the same time wanting more, even immaterial," is the main theme of the story with characters who have always stirred the human imagination, such as the moon, the sea, pirates. the latter are not only a symbol of greed for material things, but, according to history, also of people on the fringes of society, persecuted for various reasons. due to the absence of lyrics, music, authored by iztok drabik jug, who also used the electric guitar, plays an important role in creating the atmosphere. actresses and animators are barbara jamšek and elena volpi. for them, in addition to learning about the game of shadows and the use of lights, it was a great challenge to play on all sides, as they are surrounded by the audience in the show. since this is not a classic shadow theater performance where the animators are hidden behind a screen, there is a lot of emphasis on the choreography and movement. they had some problems with the acoustics in creating the show, as the church, which otherwise borders the puppet theater and was renovated last year with european funds, lacks technical equipment. according to the director of the mojca theater, they rarely looked for such an ambient performance in order to be able to take advantage of the givens of a sacral building and at the same time test the working conditions in it. otherwise, they are still waiting for the municipal tender to fill their space with a new content. |
| **Human translation of the Slovene reference summary** |
| march brings to the maribor puppet theater the premiere of the play pirate and the moon, directed by tina grabnar. a shadow theater devoted to the relationship between the material and the immaterial was placed in a minority church, with the church nave serving as a vessel. |
| **Human translation of the generated summary from Slovene**, ROUGE-L = 9,30 |
| in the play theater ljubljana ) the dennis haseleye's play about a pirate, which is based on a haseleye picture book, will premiere at 8 pm |
| **The generated summary in Slovene** |
| v predstava teatru ljubljana ) bodo drevi ob 20. uri premierno uprizorili predstavo dennisa haseleyeja o piratu , ki je nastala po motivih slikanice haseleyeja |

**Table 9** The second example (misleading) of a summary produced by the best cross-lingual summarizer.

The model can be improved in several ways. The quality of the cross-lingual alignment between Slovene and English embeddings is lower than for some other language pairs and could be improved with additional anchor points, such as bilingual dictionary. Recently introduced contextual embeddings such as BERT [13] or ELMo [36] have improved many tasks where they were applied. It would be worth testing their ability in a generative cross-lingual task such as cross-lingual summarization. Further, it may be necessary to increase the vocabulary size because of the rich Slovene morphology. Instead of ROUGE reward, RL step could maximize BERTScore reward. Instead of the two used readability measures (the internal loss function and Slovene language model) used in the selection of the generated sum-maries, we could use the recently introduced supervised or unsupervised multilingual readability approach of Martinc et al. [26]. We could improve the quality of the fine-tuning dataset by procuring news articles with the original summary-text splits (instead of the currently used heuristics). Additionally, we could denoise the Slovene dataset by calculating BERTScore scores between reference summaries (i.e. leads) and news article text and retain only the best-matching pairs.

---

[3] https://github.com/azagsam/cross-lingual-summarization

| |
|---|
| **Human translation of the original Slovene article** |
| formis: desy vahen 14, sešel 5, bračko, pintarič 20, stavbar, polanec, geratič, gajser 15, petranović 2, štern 2. kema puconci: praprotnik flisar 6, vidonja 12, vinkovič, franko 8, škraban 12, banfi 2, donša 4, jerič, koler, frumen 9, sakovič 17, kadiš 2. the volleyball players of the formis suffered another defeat, and the Kome puconci were looking forward to a new victory. rogožanke with the point won did not take a step towards the middle of the scale, so prekmurje with two points remained in its upper half. the introductory set passed in a draw, and in the playoffs the guests showed a more mature game than the hosts and took the lead with 1: 0. also in the second set, no team gained a significant advantage, rogožanke improved their game in attack and defense, and after the lead of 24:23 and the result of 25:25, they won two more points and equalized the result in the sets. the third set was again won by prekmurje, who took advantage of the poor initial blows of the hosts, and in the final they were concentrated enough not to allow a turn. the home team started the fourth set very summery and led all the time. with the result of 16:14, they dominated the field, scored points as if on a conveyor belt and tied the score at 2: 2. the decisive set was started much better by the guests, who took the lead with 5: 1 and 8: 5, then the hosts restored the balance on the field, and with the result of 9: 9, due to errors in reception and attack, three points and also win. |
| **Human translation of the Slovene reference summary** |
| the volleyball players of kema puconci defeated the home team with 3: 2 (21, - 25, 21, - 16, 15) in the match of the 7th round 1 .dol for women in hoče. * sports hall in Hoče, 130 spectators, judges: valentar (straight) and štumfelj (mežica). |
| **Human translation of the generated summary from Slovene**, ROUGE-L = 40,00 |
| volleyball players keme puconci beat formis in the 3rd round 1 dol for women with 1: 0 (1: 0 * sports hall, spectators 250, judges: bračko (kranj, štern volleyball players kema puconci are in the 2nd round 1 |
| **The generated summary in Slovene** |
| odbojkarice keme puconci so v 3. krogu 1 dol za ženske v gosteh premagale formis z 1:0 ( 1:0 * športna dvorana , gledalcev 250 , sodnika : bračko ( kranj , štern odbojkarice kema puconci so v 2. krogu 1 |

**Table 10** The third example (misrepresented numbers) of a summary produced by the best cross-lingual summarizer.

Future studies could investigate how to improve metrics for the abstractive text summarization. One idea is to combine the content-based metrics (ROUGE, BERTScore) with the perplexity measure to ensure both accuracy and readability in the same metric. An interesting problem for future work is how to attain greater levels of abstraction. In cross-lingual and model transfer research, the influence of special tokens should be studied.

**Acknowledgements** The research was supported by the Slovene Research Agency through research core funding no. P6-0411 and project no. J6-2581. The research was financially supported by European social fund and Republic of Slovenia, Ministry of Education, Science and Sport through projects Quality of Slovene textbooks (KaUč) and Ministry of Culture of Republic of Slovenia through project Development of Slovene in Digital Environment (RSDO). This paper is supported by European Union's Horizon 2020 Programme project EMBEDDIA (Cross-Lingual Embeddings for Less-Represented Languages in European News Media, grant no. 825153).